\documentclass[runningheads]{llncs}
\usepackage[T1]{fontenc}
\usepackage{graphicx}
\usepackage{booktabs}
\usepackage[misc]{ifsym}
\usepackage{times}
\usepackage{soul}
\usepackage{url}
\usepackage[hidelinks]{hyperref}
\usepackage[utf8]{inputenc}
\usepackage[small]{caption}
\usepackage{graphicx}
\usepackage{amsmath}
\usepackage{booktabs}
\usepackage{algorithm}
\usepackage{algorithmic}
\usepackage[switch]{lineno}
\usepackage{enumitem}
\usepackage{adjustbox}
\usepackage{xspace}
\usepackage[flushleft]{threeparttable}
\usepackage{subcaption}
\usepackage{float}
\usepackage{spverbatim}
\usepackage{fancyvrb}
\usepackage[margin=1.25in]{geometry}
\usepackage{xspace}
\usepackage{tikz}
\usepackage[table]{xcolor}
\usepackage{listings}
\usepackage{csvsimple}
 \usepackage{booktabs}
\definecolor{txtgray}{gray}{0.3}
\definecolor{tblgray}{gray}{0.85}

\newcommand{\Mean}{\ensuremath{\mu}}
\newcommand{\Std}{\ensuremath{\sigma}}

\usepackage{mwe}

\newif\ifAnonymous
\Anonymousfalse

\begin{document}

\title{Can LLMs Infer Conversational Agent Users' Personality Traits from Chat History?}

\titlerunning{Inferring Conversational Agent Users' Personality Traits}
\ifAnonymous\author{Anonymous~\anonymousplaceholder}\else

\author{Derya Cögendez \and
Verena Zimmermann \and
Noé Zufferey}

\authorrunning{D. Cögendez et al.}

\institute{ETH Zurich, Switzerland}
\fi

\maketitle

\begin{abstract}
Sensitive information, such as knowledge about an individual's personality, can be can be misused to influence behavior (e.g., via personalized messaging). To assess to what extent an individual's personality can be inferred from user interactions with LLM-based conversational agents (CAs), we analyze and quantify related privacy risks of using CAs. We collected actual ChatGPT logs from N=668 participants, containing 62,090 individual chats, and report statistics about the different types of shared data and use cases. We fine-tuned RoBERTa-base text classification models to infer personality traits from CA interactions. The findings show that these models achieve trait inference with accuracy (ternary classification) better than random in multiple cases. For example, for extraversion, accuracy improves by $+44\%$ relative to the baseline on interactions for relationships and personal reflection. This research highlights how interactions with CAs pose privacy risks and provides fine-grained insights into the level of risk associated with different types of interactions.
\keywords{LLM  \and privacy \and personality}
\end{abstract}

\section{Introduction}
With the increased performance of large language models (LLMs), conversational agents (CAs), such as ChatGPT, are nowadays available to any individual requiring little technical knowledge and skills. However, such tools may also cause privacy issues for the individuals who use them and eventually lead to negative consequences related to the misuse of personal information for society as a whole, e.g., through being able to personally target and manipulate people on a large scale. Previous work has shown that users often share large amounts of data with CAs, including sensitive personal data, such as health-related information~\cite{zufferey_ai_2025}. Such data is not only sensitive for its own nature and potential related malicious use cases (e.g., manipulation, discrimination), it might also be used to gain even more knowledge about specific individuals through leveraging inference models. Whereas multiple studies have analyzed how LLMs can be used to infer diverse types of personal information from text-based content~\cite{staab_beyond_2023,peters_large_2024}, no studies focused on the inference of individuals’ personal information directly from their interaction with CAs. However, such an analysis is essential to better understand and quantify the risk raised by CA usage in order to mitigate potential misuse, e.g., for manipulation and misinformation campaigns. Prior work already showed how LLMs could be used to generate personalized content to influence individuals, for example, how personality traits can be leveraged to make CAs more convincing and improve their chances of persuading users (e.g., for targeted marketing~\cite{matz_potential_2024}). Prior research in psychology showed that previous knowledge about someone's personality is highly valuable information that could be leveraged to increase persuasion likelihood and message (e.g., ads) impact~\cite{hirsh_personalized_2012,alkis_impact_2015}. Therefore, personality traits indeed provide particularly sensitive information that can be used by malicious entities for large-scale manipulation. For example, the Cambridge Analytica scandal~\cite{gibney_scant_2018} revealed that a personality-based targeted campaign on online social networks highly influenced the 2016 US presidential election. In this work, we focus on how ChatGPT users' input can be used to fine-tune a language model for personality-traits inference and the related risks for privacy. 
In particular, our contribution to this work is the following: 
(1) We report a data collection campaign of actual ChatGPT conversation logs from N=668 participants, containing 62,090 individual chats, and provide a detailed analysis of shared data types and use cases, and 
(2) We report how we used this dataset, along with a user survey that serves as our ground truth, to fine-tune and evaluate RoBERTa-base text classification models to infer personality traits from these logs. In particular, we provide fine-grained analysis of accuracies for ternary classification based on general inference, as well as types and use case-oriented inference. First, this approach allows us to guarantee unbiased results as we ensure the fine-tuned model has no previous knowledge of either the dataset or the ground truth. Second, we discuss the most sensitive data types and related misuse cases. Such information is essential to developing privacy-enhancing technologies that would mitigate privacy risks.

\section{Background - Personality Traits}
The assessment of an individual’s personality is generally based on the Big Five personality traits~\cite{mccrae_neopi3_2005}. They are also known as the five-factor model, or OCEAN model, an acronym for the five personality traits openness, conscientiousness, extraversion, agreeableness, and neuroticism.
This model has been proven to be robust and stable over time~\cite{cobb-clark_stability_2012}. The five traits can be described as follows~\cite{roccas_big_2002}:
\begin{itemize}
    \item \textbf{Openness (to experience)} High scores relate to intellectual, imaginative, sensitive, and open-minded individuals; low scores to down-to-earth, insensitive, and conventional ones.\\
    \item \textbf{Conscientiousness} High scores relate to careful, thorough, responsible, organized, and scrupulous individuals; low scores to irresponsible, disorganized, and unscrupulous ones\\
    \item \textbf{Extraversion} High scores relate to sociable, talkative, assertive, and active individuals; low scores to retiring, reserved, and cautious ones.\\
    \item \textbf{Agreeableness} High scores relate to good-natured, compliant, modest, gentle, and cooperative individuals; low scores to irritable, ruthless, suspicious, and inflexible ones.\\
    \item \textbf{Neuroticism} High scores relate to anxious, depressed, angry, and insecure individuals; low scores to calm, poised, and emotionally stable ones.
\end{itemize}
Personality traits consist of highly sensitive information by nature as highly related to individuals thinking and behavior. We provide more explanation about how such information could be leveraged to influence users in Section~\ref{section:threat}.
\section{Related Work}
Prior research showed that individuals' personality traits, and especially the Big Five traits, can be inferred from multiple data types. Previous work focused on personality traits inference from location-based data~\cite{chorley_personality_2015}, social media profile information~\cite{lima_multi-label_2014,kosinski_manifestations_2014}, text-based posts~\cite{minamikawa_blog_2011}, dietary habits~\cite{weston_personality_2020}, call details records~\cite{monsted_phone-based_2018}, smartphone usage patterns~\cite{stachl_predicting_2020}, or activity tracker's data~\cite{zufferey_watch_2023}.

Whereas most of the previous work is based on more classical machine learning approaches, more recent work also showed the potential of using language models to infer personal information from text-based content such as location, income, and gender~\cite{staab_beyond_2023}.

Inference of personality traits based on text content has also been explored in the last few years. In their article, El-Demerdash et al.~\cite{demerdash_deep_2022} describe a BERT-based approach to proceed with binary classification of personality traits and evaluated their model on two commonly-used benchmark databases for personality inference (i.e., written essays, and Facebook data), Zhu et al.~\cite{zhu_lexical_2022} proceeded with similar experiments (BERT-based binary classification) but also tested their method on two additional public datasets (i.e., Youtube speech transcriptions, Tweets). Molchanova~\cite{rapp_exploring_2024} only explored how LLM-based models can be used to binary classify users regarding their level of extraversion based on written essays. As for them, Peter et al., explored how LLMs can be used to infer social networks users personality traits based on their online activity. Most of the previous work is based on classification and achieves performance between ($+20\%$) and ($60\%$) better than the random baseline, depending on the model, data source, and inferred trait. Finally, Wright et al. explored the correlation between personality traits and zero-shot inference based on transcription of recorded free-speech~\cite{wright_assessing_2026}, and showed that extraversion and neuroticism are particularly sensitive traits in such context.

As such, we can assume that users are prone to privacy inference risks related to the widespread use of CAs, given that sensitive data that is often shared in the conversations \cite{zufferey_ai_2025,malki_hoovered_2025}. However, to our knowledge, no prior work has analyzed how personality traits can be inferred from interaction logs with CAs. Such a study would indeed provide more specific insight toward privacy issues related to the usage of LLM-base CAs. Our article fills this gap, thereby providing quantification of privacy risks related to CA usage and valuable insights that can inform effective countermeasures that protect the users' privacy.
\section{Adversary Model}
\label{section:threat}
We mainly consider adversaries such as LLM-based service providers and their business partners. Indeed, as service providers are in charge of managing the storage of the data, they are the most likely to access it. Such a model could also be extended to service provider partners, such as their model/API provider and cloud-storage provider. In a worst-case scenario, logs could even be sold to data brokers. Aside from an honest-but-curious adversary (that would just gain knowledge about their users attributes), we also consider the more malicious cases where the service provider likes to leverage users' personality traits for more influence/manipulation, such as targeted advertisement or even political propaganda. For example, OpenAI has announced that it will start testing advertisements for US-based ChatGPT users~\cite{openai2026_advertising}. In such cases, they could consider leveraging users' personal information to optimize B2B services. Besides service providers, other malicious entities could access CA user data in multiple ways. For example, in summer 2025, a massive amount of Grok and ChatGPT conversation logs have been leaked and were even directly available from Google search~\cite{mcmahon_hundreds_2025,sims_thousands_2025}, both due to poor user interface design, misleading users to unwillingly publicly share their conversations. Other apps, such as VPN services~\cite{dardikman_8_2025}, also contain hidden eavesdropping scripts that simply extract conversations. All previously described adversary could then used acquired knowledge to, first, gain personal information that have been directly shared to CAs by users, and, second, use such data to infer other highly sensitive one, as personality traits, who are known to be highly efficient and used in the past for individuals' manipulation and large-scale influence as for determining if a particular individual would likely reimburse personal debts after a reassuring or threatening message~\cite{duhigg_what_2009}, and to craft targeted messages in the context of an national election campaign on online social network~\cite{gibney_scant_2018}. This, therefore consists a major threat, especially knowing that personality-trait-related information is also efficient for automatized LLM-powered persuasion~\cite{matz_potential_2024}.
\section{Methodology}
In the following, we describe our approach to the data collection and analysis. Please see the technical appendix for more details about the dataset, used prompts, and training parameters.

\subsection{Data Collection}
We recruited individuals to participate in our research through the platforms Prolific and Clickworker.
The participants were exclusively from the US and UK and were fluent in English.
Our survey included a standard scale questionnaire to assess their personality traits, and questions on basic demographics (e.g., gender, age).
During the survey, participants were also asked to upload their ChatGPT history to the platform.
Data collection resulted in 668 individuals who passed attention checks and uploaded valid interaction logs, with 62,090 individual chat sessions.

\subsection{Ethical and Open Science Statement}
Being aware of the potential privacy concerns related to user interaction with CAs, we took the greatest care to conduct the research in an ethical and privacy-preserving way. Our work followed standards for ethical-psychological research and has been reviewed and approved by the IRB of our institution. 
The compensation followed the platform's recommendation for fair payment for a study duration of approximately 30 minutes (20 minutes questionnaire and 10 minutes upload of CA conversation history).
Each participant was paid £7.70 or €8.7 ($\sim$\$10).
A technical appendix with the dataset, analysis, and training details is submitted with this article.
However, due to the highly sensitive nature of its content, the dataset can unfortunately not be released.
Furthermore, participants were never asked to share data with CAs for the purpose of the study, but we relied on previous logs of conversations that users had voluntarily had with CAs outside this study. Participants were informed in detail about the nature and purpose of this research.
As the data collection involved the collection of demographic information, personality traits, and potentially sensitive logs of the interaction with ChatGPT, we took multiple steps to protect the participants' privacy. In particular, uploads were encrypted, data was only stored on secure servers of our institution, and was only accessible to the involved researchers within our institution's secure network (or through VPN). Furthermore, data was only analyzed at the level of data types disclosed in the chat logs, and we removed all account-related files containing usernames and emails. Any use and processing of the data (e.g., log analysis, model training) was conducted on a secure virtual machine from our institution. We did not use API, cloud storage, or external services of any kind. We committed to delete all collected data within a maximum of five years. For scientific publications, only aggregated and properly de-identified data will be reported.

\subsection{Dataset}
To build an inference model for personality traits, we needed to collect the ground truth for each participant. We chose to rely on the Big Five model~\cite{mccrae_neopi3_2005} that defines an individual's personality through five main traits. We assessed the ground truth for personality traits of each respondent based on their answers to a standardized 60-item evaluation scale (IPIP-60) that was included in our user survey~\cite{maples-keller_using_2019}. For each trait, each user's IPIP-60 scores were then categorized as low, medium, or high based on terciles computed from the collected dataset (simulating classes based on a global population), each class therefore contains one third of our participants for each trait. We then evaluate personality inference using ternary classification accuracy using these labels.

Each user chat log was treated as a single data point, with five labels corresponding to the user’s personality traits. We only used text-based user interactions and excluded all model-generated responses. As a result, the model input consists of user-generated messages along with the associated chat titles.

The length of user chat histories varied substantially. On average, users in our dataset had 92 interactions with ChatGPT, while the median history length was 28 chats (std = 185.72). This indicates that many users have relatively short histories, and a smaller number have very large ones. For instance, the largest history in the dataset contained over 2000 text-only interactions.

\paragraph{Statistics.}
The survey respondents were $54\%$ men, $45\%$ women, four non‑binary individuals $(<1\%)$, and three persons who did not disclose their gender $(<1\%)$.
Distributions of UK and US populations were $51\%$ and $49\%$ respectively. The personality trait distribution of our UK sample was similar to that of the general UK population, as reported in a study of approximately 386,000 individuals \cite{rentfrow_regional_2015}. 
The largest difference between the study and our dataset was seen in the openness trait, with the mean scores of $3.67 \pm 0.64$ and $3.37 \pm 0.49$, respectively, on a Likert scale. Distribution of Big-Five scores between the UK and USA populations does not differ substantially, except for a slightly higher overall agreeableness in the USA \cite{schmitt_geogragphic_2007}.
We observe this trend for agreeableness in our dataset.
However, lower bound thresholds for agreeableness labels did not differ significantly (UK: $0$, $3.5$, $4.0$, US: $0$, $3.75$, $4.17$), so we combine these samples.

\subsection{User Chat Analysis}
To analyze the dataset further, we label the user chats in two dimensions: types of personal data shared and use intents in their interactions with ChatGPT. To generate these labels, we chose Qwen3-8B because of its general reasoning performance relative to similarly sized models~\cite{yang2025qwen3technicalreport}. Due to computational constraints, we use the AWQ 4-bit version.

For each dimension, the model is provided with a user chat and a list of labels and is asked to select the most relevant label based on the user’s inputs. We then extract the corresponding labels with text processing. When multiple labels are relevant, the model is instructed to choose a single label. The user chats are truncated to 1,000 tokens because during initial testing, we found that longer inputs caused context issues, with the model following instructions from the user chat rather than the labeling task. 

Labeling of types of personal data shared was done based on previous work about self-declared personal data sharing habits with LLMs, where the authors describe personal data in four main categories (such as \textit{Lifestyle and Health} and \textit{Personal Characteristics and Emotions}) \cite{zufferey_ai_2025}. This process helped us to calculate statistics about actual users' personal data sharing practices. 

User intent labeling was based on ChatGPT use cases identified by OpenAI in the working paper \textit{How People Use ChatGPT} \cite{NBERw34255}. These use cases were grouped into seven main categories such as \textit{Writing}, \textit{Technical Help}, and \textit{Seeking Information}. While the original study labels individual user messages, we used the same labels for the entire chat. We also used a condensed version of their prompt.

\subsection{Personality Traits Inference}

We simulated ternary classification attacks for each personality trait. In other words, the attacker's goal is to guess, for each user, based on their chat history content if they scores low, medium, or high in each of the five personality traits. 
We proceeded to classification because (1) in terms of personality traits, the category which an individual belongs to is a very important aspect, and personality tests are also using quantile-based (or z-score based) evaluation rather than only raw-score (i.e., compared to the exact score)~\cite{mccrae_neopi3_2005} and (2) it is one the most common method used in previous related work~\cite{zufferey_watch_2023,monsted_phone-based_2018,hassanein_predicting_2021}. It is therefore likely that an adversary, as defined in \autoref{section:threat} will base their attack on a similar method. Furthermore, relying on classification, as well as accuracy as main evaluation metric, allows us to depict particularly clear quantification of the related privacy threat, giving one easily understandable metric, i.e., the proportion of users that are correctly classified. Such an approach allows for effective science communication and public privacy awareness.
Classes can be defined based on quantiles in order to get evenly sized groups (in terms of their number of individuals).
As in previous work with ternary classification of personality traits~\cite{zufferey_watch_2023,monsted_phone-based_2018}, the classes are defined using the collected dataset terciles for each trait.
Therefore, we evaluate the attack on evenly distributed classes. Based on this adversary's goal, we simulated personality inference attacks in five different variations:

\begin{enumerate}
    \item zero-shot inference for each user
    \item trained general classification for each chat
    \item trained general classification for each user
    \item trained content-based classification for each user (data-type based)
    \item trained content-based classification for each user (use-case based)
\end{enumerate}

\subsubsection{Zero-shot Inference.}

As previous work showed the capacity of LLMs to be used for zero-shot inference of personality traits from text~\cite{wright_assessing_2026}, we decided to first simulate such an attack.
This inference attack consists of, for each of the N=668 users, prompting an LLM to fill in the Ten Item Personality Inventory (TIPI) based on their input~\cite{gosling_very_2003}. The TIPI is a 10-item short version Big Five personality assessment scale, containing two questions per personality trait, asking the respondent (here substituted by the LLM) to rate how much they agree with the following statements: ``I see myself as:'' 
\begin{itemize}[noitemsep]
    \item Open to new experiences, complex. (Openness)
    \item Conventional, uncreative. (Openness)
    \item Disorganized, careless. (Conscientiousness)
    \item Dependable, self-disciplined. (Conscientiousness)
    \item Extraverted, enthusiastic. (Extraversion)
    \item Reserved, quiet. (Extraversion)
    \item Critical, quarrelsome. (Agreeableness)
    \item Sympathetic, warm.  (Agreeableness)
    \item Anxious, easily upset. (Neuroticism)
    \item Calm, emotionally stable. (Neuroticism)
\end{itemize}

For this attack, we used the model Qwen3 8B AWQ (4-bit quantization version of Qwen3 8B) from Qwen AI, which is used for zero-shot inference. We chose this model for its reasoning capabilities, which are required for this task, and as Qwen models have been proven effective in such inference in previous related works~\cite{wright_assessing_2026}. The model is prompted to agree or disagree with each statement for each data point, using chain-of-thought prompting. This process uses the following prompt:

{\footnotesize
\begin{Verbatim}[frame=single]
### **Task Overview**

Imagine you are the person who wrote the following user messages.
Read them all and internalize their tone, style, and behavior.
Then, as this person, complete the Ten-Item Personality Inventory (TIPI)
by answering each item from 1 (Disagree strongly) to 7 (Agree strongly),
based on how you - as that person - see yourself.

**User Messages to Analyze**

{input}

---

### **Ten-Item Personality Inventory (TIPI)**

Rate the following items from 1 (Disagree strongly) to 7 (Agree strongly).

1. Extraverted, enthusiastic.
2. Critical, quarrelsome.
3. Dependable, self-disciplined.
4. Anxious, easily upset.
5. Open to new experiences, complex.
6. Reserved, quiet.
7. Sympathetic, warm.
8. Disorganized, careless.
9. Calm, emotionally stable.
10. Conventional, uncreative.

Note: This is a speculative exercise. The goal is not to produce a perfectly
accurate personality profile, but rather a reasonable approximation based on
patterns in the user’s language and behavior. Answer each item as if you were
the person who wrote the messages, using your best judgment.

Imagine you are the person who wrote these messages and complete the 10 item
form by selecting a number from 1 to 7 for each item above, based on how you
believe this person sees themselves.
\end{Verbatim}
}

\begin{figure}[tbp]
     \centering
     \includegraphics[width=.75\columnwidth]{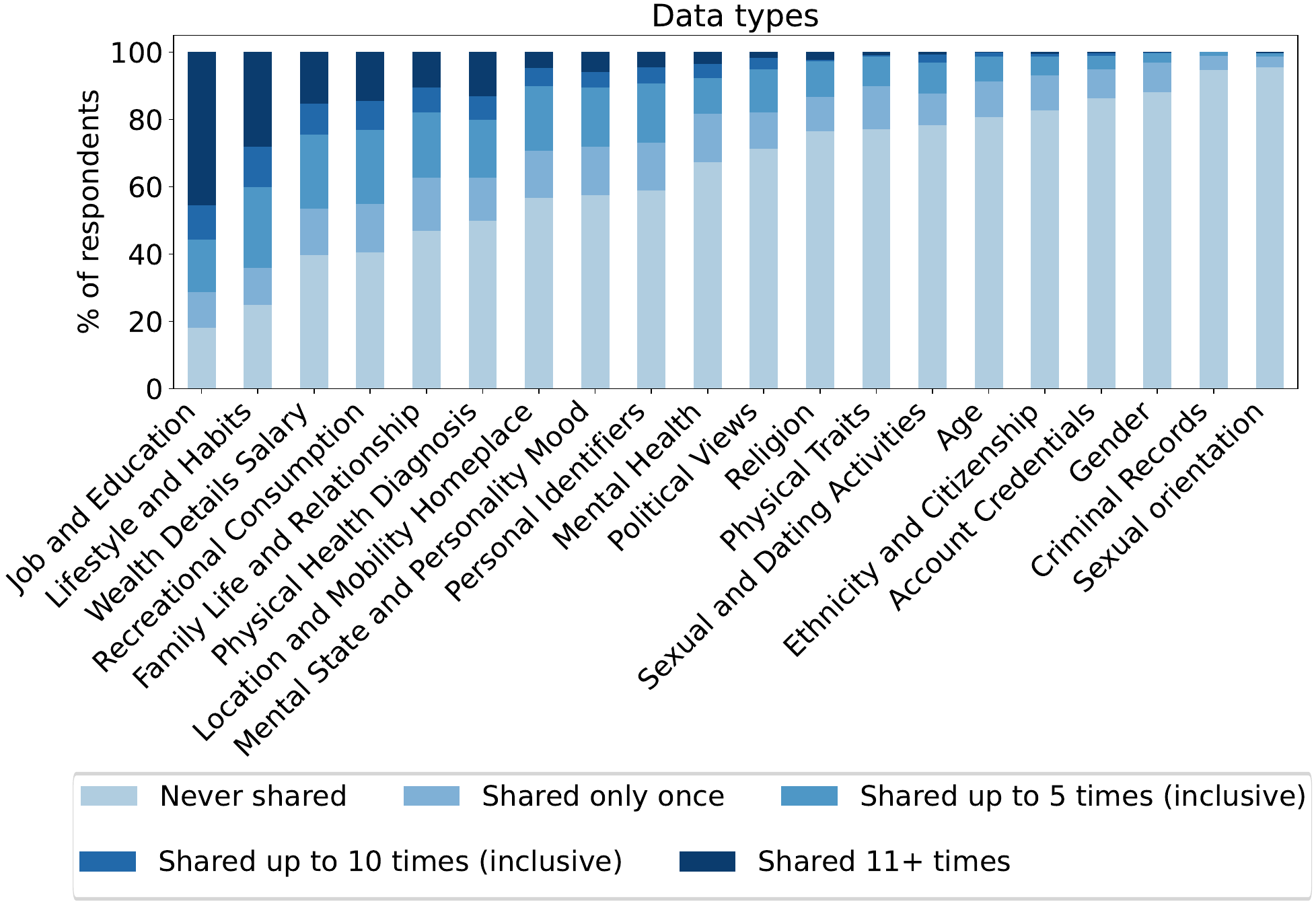}
     \caption{Number of times participants shared specific personal information with ChatGPT (N=668).}
     \label{fig:share_type}
 \end{figure}
Then, we extract the LLM's ratings from the output text for each item on the TIPI test, and use the final score to predict the score for each personality trait as low, medium, or high. 
For technical reasons, we had to limit all the used user inputs to a size of 1000 tokens. We split all inputs corresponding to each user into multiple text chunks and proceeded with the zero-shot inference for each of them. Finally, we proceeded to a majority-vote-based classification, i.e., we selected the inferred class that was predicted most frequently for each user.

\subsubsection{Fine-tuned Classifier.}
To go beyond Zero-shot inference, and as, according to our adversarial model, we can assume adversaries having access to sufficient resources, we also decided to evaluate to what extent a fine-tuned model would perform. We then fined-tuned five (i.e., for each trait) RoBERTa-base \cite{zhuang-etal-2021-robustly} models for personality trait prediction using five-fold cross-validation. While constructing the folds, we ensured that each survey participant appeared in only one fold to prevent test-leakage, following the ``dos and don'ts'' of machine learning~\cite{arp_dos_2020}. Because chat history lengths varied significantly across users, this split led to uneven label distributions. We sub-sampled within each fold to ensure uniform training and test sets, and we repeated this process for each personality trait.

For each of the classifiers, we performed grid search for the hyperparameter selections and chose the final parameters based on the mean evaluation accuracy across the five folds. We report test accuracy using both individual chat classifications and predictions per user aggregated across chats. To compute a single prediction per user, we assigned each user the most frequently predicted label among their chats. In cases where multiple labels were tied, we randomly selected one of the tied labels as the prediction.

Finally, we evaluated how the previously model performs when the inference is only based on specific chats regarding two different chat classification (1) what types of personal data are included in the prompts, and (2) what is the general use case. Evaluation of inference performance regarding conversations containing particular data types or use cases offers a more fine-grained analysis of the risks. In particular, it allows us to identify the types of data or usage patterns that most likely enhance the accuracy of inference models.

For the evaluation of all fine-tuned-based inference attack, we proceeded to 5-fold cross-validation in order to increase confidence level. For each inference model, we so randomly separated the dataset, making sure that each user only appears in one fold (to avoid data-leakage), and making sure the label distributions are approximately uniform. However, this latter step requires randomly removing some data points, resulting in a slightly different number of users and chats in the evaluation process. In the results sections, we indicate each of these numbers for all results.

  \begin{figure}[tbp]
     \centering
     \includegraphics[width=.75\columnwidth]{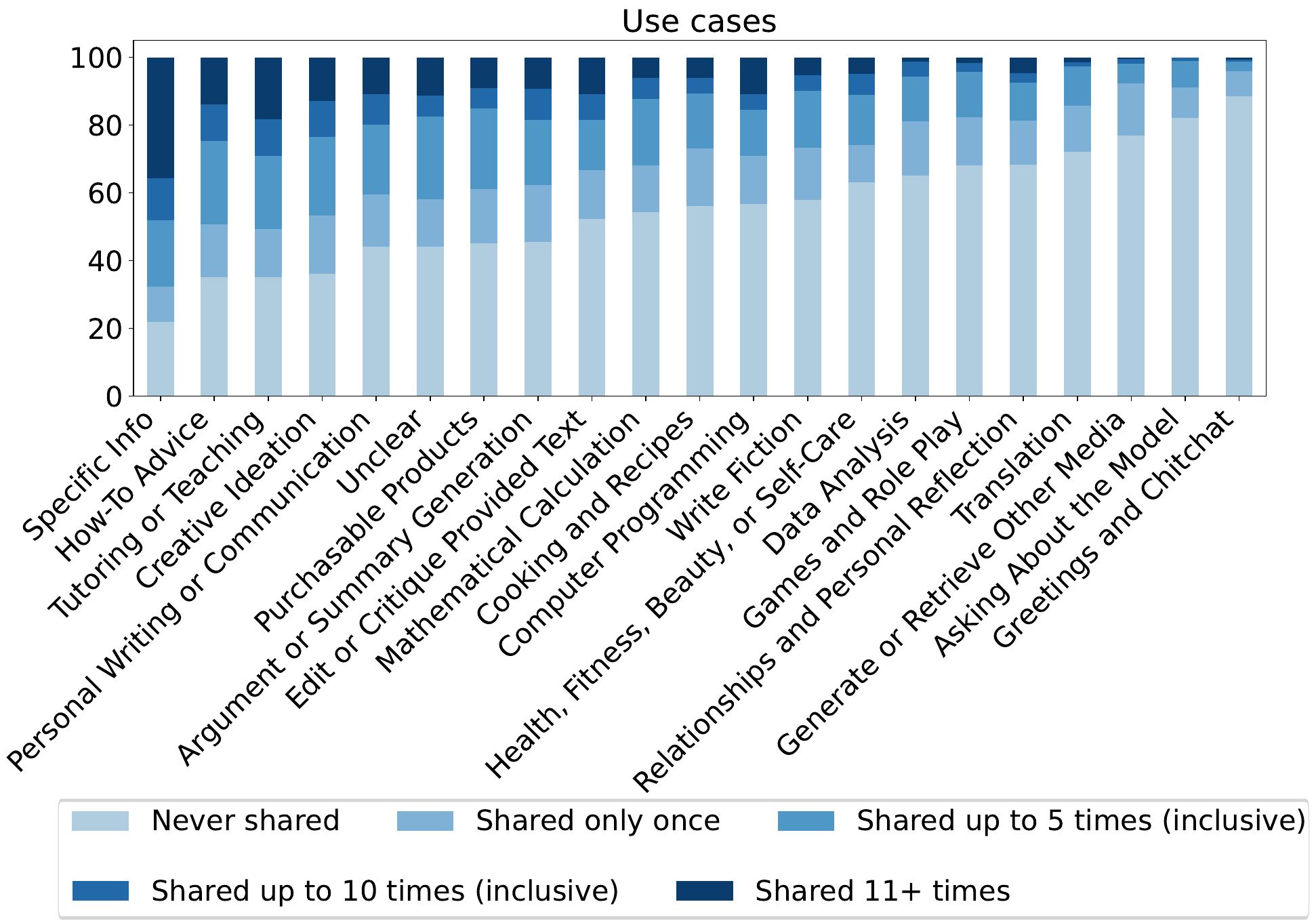}
     \caption{Number of times participants interacted with ChatGPT for specific use cases (N=668).}
     \label{fig:share_use}
 \end{figure}
\section{Results}
We first describe what types of personal data are shared when using CAs and how often. Second, we analyze the main use cases. Both aspects allow us to better understand the links between usage behavior and privacy risks resulting from inference. The first one is also a major contribution to better understanding users and privacy risks related to direct sharing of (potentially sensitive) personal data. We then move on to personality inference using these analyses.
\subsection{Labeling User Chats}
\paragraph{Personal Data Shared}

\autoref{fig:share_type} shows the percentage of users sharing specific data types sorted by how frequently they shared this data type in their chat logs. Our results generally confirm most of self-declared behaviors as described in prior work. For example, our data shows that at least half of the users shared data related to \textit{Job and Education}, \textit{Lifestyle and habits}, \textit{Family Life and Relationship}, and \textit{Physical Health} at least once, which corresponds to previous work~\cite{zufferey_ai_2025}. Yet, our data shows that \textit{Wealth Details and Salary} and \textit{Recreational Consumption} are shared more often than users declare. Some findings even contradict previous work, especially for gender, age, physical traits, and sexual orientation, which were expected to be shared more often based on prior work showing that users generally share more data in practice than they declare they are willing to disclose~\cite{malki_hoovered_2025}.

\begin{table}
\centering
\caption{Inference accuracy across all chats for each trait. Inference models are not statistically better than the random baseline (McNemar test).}
\label{table:acc_vote_zero}
\begin{threeparttable}
\rowcolors{1}{white}{tblgray}
 \begin{tabular}{l|l|c|l|}
  {}  &{\bf Accuracy}	&{\bf P-value}	&{\bf N}\\
  \hline
{\bf O}	&39.8 (+19.5\%)	& ns &668\\
{\bf C}	&36.4 (+9.1\%)	& ns &668\\
{\bf E}	&35.8 (+7.3\%)	& ns &668\\
{\bf A}	&36.4 (+9.1\%)	& ns &668\\
{\bf N}	&36.7 (+10.0\%)	& ns &668\\
 \end{tabular}
 \end{threeparttable}
\end{table}

\paragraph{Use Cases}
\autoref{fig:share_use} shows the percentage of our participants using ChatGPT for specific use cases and how frequently users used it to achieve a related task. In line with recent findings~\cite{NBERw34255}, our results show that at least half or more of the users used it for \textit{Specific Info}, \textit{How-To Advice}, \textit{Tutoring or Teaching}, \textit{Creative Ideation}, \textit{Personal Writing or Communication}, \textit{Purchasable Products}, and \textit{Argument or Summary Generation} at least once. Whereas our results about specific use cases and directly shared data give valuable information about how users' privacy is at risk, it could also be used to infer additional personal data, such as, for example, general interests, occupation, concerns, opinions, and, in our case of interest, personality traits.

\subsection{Zero-shot inference}

In the following, we report the results for zero-shot personality traits inference for each respondent.

\autoref{table:acc_vote_zero} shows the accuracy of such a zero-shot attack. Our results suggest that a reasoning model cannot predict significantly better than the random baseline when it comes to zero-shot inference based on user prompts. Except for Openness, all accuracies are lower than those reported in \autoref{table:acc_vote}. Although previous work~\cite{wright_assessing_2026} showed the effectiveness of zero-shot methods to infer personality traits, our results demonstrates the limitation of such methods and how they might vary across context. Indeed, in their article, Wright et al. report an experiment based on so-called ``open-ended narratives'' (i.e., letting the participants speak freely for a given amount of time), a context, particularly favorable to introspection, and so collection of data insightful toward one's personality, differs from interaction with CAs. In other words, we can assume that people interact in different ways with other people, themselves, and CAs.

\begin{table}
\centering
\caption{Inference accuracy across all chats for each trait. All inference models are statistically better than the random baseline (McNemar test).}
\label{table:acc_chats}
\begin{threeparttable}
\rowcolors{1}{white}{tblgray}
 \begin{tabular}{l|l|c|l|}
  {}  &{\bf Accuracy}	&{\bf P-value}	&{\bf N}\\
  \hline
{\bf O}	&36.9 (+10.7\%)	&$p < 0.01$ &46432\\
{\bf C}	&39.1 (+17.4\%)	&$p < 0.01$ &49618\\
{\bf E}	&44.2 (+32.6\%)	&$p < 0.01$ &60121\\
{\bf A}	&40.2 (+20.5\%)	&$p < 0.01$ &58607\\
{\bf N}	&38.1 (+14.2\%)	&$p < 0.01$ &49672\\
 \end{tabular}
 \end{threeparttable}
\end{table}

\subsection{General Inference with Trained Classifier}
As zero-shot inference did not significantly perform better than random, we decided to also evaluate how a specifically fine-tuned model would perform.
In this section, we describe our results on the extent to which classifiers can infer users' personality traits.
\paragraph{Single Chat Inference.}
\autoref{table:acc_chats} depicts the mean test inference accuracy of the models trained. These models were evaluated and tested for each of the personality traits with five-fold cross-validation for personality traits inference from single chats (i.e., one inference for each logged chat session). As we can see, we achieved significantly higher accuracies than the baseline for all traits. For extraversion and agreeableness, we achieve $+32.6\%$ and $+20.5\%$ accuracy compared to random. However, for some traits, and in particular for openness, conscientiousness, and neuroticism, the model performance is still not very high, with an increase of accuracy lower than $20\%$ compared to random classification. Furthermore, as we evaluated inference from single chats, the model generally achieves multiple different inferences for one specific user~\footnote{Note that we ensured that no data from the same user was used in both training and test set in any case.}. Whereas these results provide information about a large-scale general inference attack, we may lack essential information to understand privacy risks for a single user, as each user's personality trait gets one prediction for each chat (i.e., multiple predictions~per~user).

\begin{table}
\centering
\caption{Inference accuracy across users based on majority votes for each trait. Inference models for Agreeableness and Neuroticism are statistically better than the random baseline (McNemar test).}
\label{table:acc_vote}
\begin{threeparttable}
\rowcolors{1}{white}{tblgray}
 \begin{tabular}{l|l|c|l|}
  {}  &{\bf Accuracy}	&{\bf P-value}	&{\bf N}\\
  \hline
O	&36.3 (+9.0\%)	&ns&655\\
C	&37.0 (+10.9\%)	&ns&657\\
E	&36.1 (+8.2\%)	&ns&668\\
{\bf A}	&{\bf 40.6 (+21.9\%)}	&$p < 0.05$ &667\\
{\bf N}	&\bf{41.0 (+22.9\%)}	&$p < 0.05$ &659\\
 \end{tabular}
 \end{threeparttable}
\end{table}

\begin{table*}
\centering
\small
\caption{Inference accuracy across user having shared specific data types based on majority votes for each trait. Min. and max. number of users for each type are in parentheses.}

\label{table:acc_types}
\begin{threeparttable}
\rowcolors{1}{white}{tblgray}

\begin{tabular}{l|l|l|l|l|l}
                         {\bf Data type} &                    {\bf O} &                          {\bf C} &                          {\bf E} &                    {\bf A} &                          {\bf N}\\
\midrule
              Account Credentials (78-97) &     $39.2$ $(+18\%)$ &           $42.1$ $(+26\%)$ &           $44.4$ $(+33\%)$ &      $34.1$ $(+2\%)$ &           $44.8$ $(+34\%)$\\
                              Age (120-137) &      $35.1$ $(+5\%)$ &           $37.9$ $(+14\%)$ &           $44.2$ $(+33\%)$ &     $44.7$ $(+34\%)$ &            $36.4$ $(+9\%)$\\
                 Criminal Records (30-38) &     $51.9$ $(+56\%)$ &            $34.4$ $(+3\%)$ &           $36.8$ $(+10\%)$ &    $28.4$ $(-15\%)$ &            $33.3$ $(+0\%)$\\
        Ethnicity and Citizenship (103-124) &     $39.5$ $(+18\%)$ &           $43.3$ $(+30\%)$ &            $34.5$ $(+4\%)$ &      $35.5$ $(+6\%)$ &           $42.4$ $(+27\%)$\\
     Family Life and Relation. (350-376) &      $35.9$ $(+8\%)$ &            $35.6$ $(+7\%)$ &            $36.1$ $(+8\%)$ &     $38.6$ $(+16\%)$ & ${\bf 41.4}$ ${\bf (+24\%)^{\dagger}}$\\
                           Gender (69-84) &      $33.8$ $(+1\%)$ &           $39.8$ $(+20\%)$ &           $40.5$ $(+21\%)$ &     $47.4$ $(+42\%)$ &           $41.3$ $(+24\%)$\\
                Job and Education (563-586) &     $36.9$ $(+11\%)$ &           $36.9$ $(+11\%)$ &           $37.5$ $(+12\%)$ & ${\bf 41.1}$ ${\bf (+23\%)^{*}}$ &           $38.2$ $(+15\%)$\\
             Lifestyle and Habits (505-538) &     $38.6$ $(+16\%)$ &            $35.6$ $(+7\%)$ &           $37.7$ $(+13\%)$ &     $36.8$ $(+10\%)$ & ${\bf 40.3}$ ${\bf (+21\%)^{\dagger}}$\\
  Location and Mobility (284-308) &     $36.5$ $(+10\%)$ &            $34.9$ $(+5\%)$ &            $35.6$ $(+7\%)$ &      $34.8$ $(+4\%)$ &           $37.6$ $(+13\%)$\\
                    Mental Health (210-230) &     $38.6$ $(+16\%)$ &           $38.1$ $(+14\%)$ &      ${\bf 47.0}$ ${\bf (+41\%)^{**}}$ &     $36.6$ $(+10\%)$ &           $39.4$ $(+18\%)$\\
Mental State, Pers., Mood (269-304) & ${\bf 44.0}$ ${\bf (+32\%)^{*}}$ &           $41.1$ $(+23\%)$ &           $38.5$ $(+15\%)$ &     $40.8$ $(+22\%)$ & ${\bf 42.7}$ ${\bf (+28\%)^{\dagger}}$\\
             Personal Identifiers (264-293) &      $34.8$ $(+4\%)$ &           $37.0$ $(+11\%)$ &           $38.4$ $(+15\%)$ &     $40.3$ $(+21\%)$ &           $41.5$ $(+25\%)$\\
        Physical Health Diagnosis (331-359) &      $34.6$ $(+4\%)$ &            $35.6$ $(+7\%)$ & ${\bf 40.7}$ ${\bf (+22\%)^{\dagger}}$ &     $38.0$ $(+14\%)$ &           $38.6$ $(+16\%)$\\
                  Physical Traits (138-161) &     $39.0$ $(+17\%)$ &           $40.6$ $(+22\%)$ &           $44.6$ $(+34\%)$ &     $39.8$ $(+19\%)$ &           $42.0$ $(+26\%)$\\
                  Political Views (182-200) &     $41.9$ $(+26\%)$ &           $37.7$ $(+13\%)$ &           $42.9$ $(+29\%)$ &     $40.1$ $(+20\%)$ &           $40.8$ $(+22\%)$\\
         Recreational Consumption (397-424) &     $37.5$ $(+12\%)$ &           $36.6$ $(+10\%)$ &           $37.0$ $(+11\%)$ &      $34.9$ $(+5\%)$ & ${\bf 41.8}$ ${\bf (+25\%)^{\dagger}}$\\
                         Religion (147-165) &     $39.5$ $(+18\%)$ & ${\bf 45.1}$ ${\bf (+35\%)^{\dagger}}$ &           $40.9$ $(+23\%)$ &     $43.9$ $(+32\%)$ &            $36.0$ $(+8\%)$\\
     Sex. and Dating Activities (37-155) &     $37.6$ $(+13\%)$ &           $39.6$ $(+19\%)$ &           $44.1$ $(+32\%)$ &     $38.6$ $(+16\%)$ &           $31.9$ $(-4\%)$\\
               Sexual orientation (23-33) &     $41.0$ $(+23\%)$ &           $50.0$ $(+50\%)$ &           $63.3$ $(+90\%)$ &    $24.2$ $(-27\%)$ &           $38.0$ $(+14\%)$\\
            Wealth Details Salary (398-430) &     $40.0$ $(+20\%)$ &            $33.2$ $(+0\%)$ &           $37.7$ $(+13\%)$ &     $38.6$ $(+16\%)$ & ${\bf 41.0}$ ${\bf (+23\%)^{\dagger}}$\\
\end{tabular}
  \begin{tablenotes}
    \item We used a Mcnemar test to compare model performance with the random classification baseline. Inferrences achieving a p-value $< 0.1$ are in bold and annotated as follow: $^{\dagger}p < 0.1$, *$p < 0.05$, **$p < 0.01$.
   \end{tablenotes}
 \end{threeparttable}
\end{table*}

\paragraph{Majority Vote for Single User Inference.}
To evaluate global risks for a single user, we proceeded to a majority-vote-based inference model, i.e., based on the single-chat previously described results, we selected the inference class that was predicted most frequently for each user. \autoref{table:acc_vote} depicts the related results. As we can see, except for neuroticism, the general inference accuracy is almost unchanged or slightly drops. This is probably due to different usage patterns across users. However, our results show that inference models achieve performance significantly better than random for agreeableness and neuroticism, with respectively $+21.9\%$ and $+22.9\%$ accuracy compared to random guess.

\subsection{Content-based Inference Models}
Evaluation of inference performance regarding conversations containing particular data types or use cases offers a more fine-grained analysis of the risks. In particular, it allows us to identify the types of data or usage patterns that most likely enhance the accuracy of inference models.
\autoref{table:acc_types} depicts inference accuracies for each personality trait and data type. We can see that inference based on data types achieves significant ($p<0.05$) or weak evidence ($p<0.1$) results for all traits. In particular, information related to \textit{Mental Health} is very informative to infer extraversion ($+41\%$). Specifically looking at the other significant results (i.e., p<0.05), our results also show that data related to \textit{Mental State, Personality, and Mood} is very informative for openness inference ($+32\%$), and information related to \textit{Job and Education} is particularly informative to infer accuracy ($+23\%$). Beyond significance, our results also suggest that users sharing information related to their \textit{Sexual orientation} are particularly sensitive to the inference of extraversion ($+90\%$), and conscientiousness ($+50\%$) level, however, we have too few related user-data to conclude with solid proof.

As for inference focused on use cases depicted in \autoref{table:acc_cases}, our results show that extraversion is particularly sensitive to several use cases, and in particular, \textit{Relationship and Personal Reflection}, with which inference model achieves $+44.0\%$ of performance, and \textit{Health, fitness, beauty, and self-Care}, with which it achieves $+32.0\%$ of performance. However, for the other traits, the use-case-based inferences are generally similar or less informative than the data-type-oriented ones for most of the traits. Indeed, whereas data types classification rather corresponds to users' characteristics (i.e., what the user is), use cases are more related to action (i.e., what the user does). This might also explain why extraversion is the most sensitive trait to inference based on use cases, as this trait is the closest one to action~\cite{mccrae_neopi3_2005}.

\begin{table*}
\centering
\small
\caption{Inference accuracy across users having specific usage of CAs based on majority votes for each trait. Min. and max. number of users for each case are in parentheses.}
\label{table:acc_cases}
\begin{threeparttable}
\rowcolors{1}{white}{tblgray}

\begin{tabular}{l|l|l|l|l|l}
                         {\bf Use case} &                    {\bf O} &                          {\bf C} &                          {\bf E} &                    {\bf A} &                          {\bf N}\\
\midrule
       Arg. or Summary Gen. (359-386) &  $36.4$ $(+9\%)$ &           $37.2$ $(+12\%)$ &           $39.1$ $(+17\%)$ &           $39.3$ $(+18\%)$ &           $40.0$ $(+20\%)$ \\
               Asking About the Model (112-125) &  $35.1$ $(+5\%)$ &           $42.1$ $(+26\%)$ &           $41.1$ $(+23\%)$ &           $42.3$ $(+27\%)$ &           $31.1$ $(-7\%)$ \\
                 Computer Programming (281-311) & $39.3$ $(+18\%)$ &           $37.4$ $(+12\%)$ &           $39.5$ $(+19\%)$ &           $38.7$ $(+16\%)$ &            $35.5$ $(+7\%)$ \\
                  Cooking and Recipes (286-311) & $40.1$ $(+20\%)$ &           $37.3$ $(+12\%)$ &           $40.3$ $(+21\%)$ &           $37.4$ $(+12\%)$ &            $34.7$ $(+4\%)$ \\
                    Creative Ideation (428-452) & $38.0$ $(+14\%)$ &            $36.5$ $(+9\%)$ &            $36.2$ $(+9\%)$ &           $37.6$ $(+13\%)$ & ${\bf 41.2}$ ${\bf (+24\%)^{\dagger}}$ \\
                        Data Analysis (218-249) & $38.8$ $(+16\%)$ &            $34.5$ $(+4\%)$ &           $37.4$ $(+12\%)$ &           $38.6$ $(+16\%)$ &            $34.9$ $(+5\%)$ \\
       Edit/Critique Provided Text (315-341) & $37.3$ $(+12\%)$ &           $40.8$ $(+22\%)$ &           $39.4$ $(+18\%)$ & ${\bf 42.1}$ ${\bf (+26\%)^{\dagger}}$ &            $34.3$ $(+3\%)$ \\
                  Games and Role Play (197-226) & $38.8$ $(+17\%)$ &           $37.6$ $(+13\%)$ &           $38.6$ $(+16\%)$ &           $38.3$ $(+15\%)$ &            $34.1$ $(+2\%)$ \\
     Gen./Retrieve Other Media (135-160) &  $35.5$ $(+7\%)$ &            $34.5$ $(+4\%)$ & ${\bf 44.7}$ ${\bf (+34\%)^{\dagger}}$ &           $41.0$ $(+23\%)$ &           $37.2$ $(+11\%)$ \\
                 Greetings and Chitchat (67-82) & $31.8$ $(-5\%)$ &            $33.8$ $(+2\%)$ &           $41.3$ $(+24\%)$ &           $43.0$ $(+29\%)$ &           $32.0$ $(-4\%)$ \\
Health/Fit./Beauty/S.-Care (242-261) &  $33.4$ $(+0\%)$ &            $35.9$ $(+8\%)$ &       ${\bf 44.0}$ ${\bf (+32\%)^{*}}$ &           $37.2$ $(+12\%)$ &           $40.8$ $(+23\%)$ \\
                        How-To Advice (438-463) & $37.8$ $(+13\%)$ &           $36.7$ $(+10\%)$ &           $36.6$ $(+10\%)$ &           $39.3$ $(+18\%)$ &            $35.8$ $(+7\%)$ \\
             Mathematical Calculation (302-327) & $37.9$ $(+14\%)$ &           $39.5$ $(+18\%)$ &           $39.6$ $(+19\%)$ &           $37.7$ $(+13\%)$ &           $37.1$ $(+11\%)$ \\
    Personal Writing or Comm. (362-396) & $39.6$ $(+19\%)$ &            $34.7$ $(+4\%)$ & ${\bf 40.6}$ ${\bf (+22\%)^{\dagger}}$ &            $36.2$ $(+9\%)$ &           $38.9$ $(+17\%)$ \\
                 Purchasable Products (360-392) &  $35.3$ $(+6\%)$ &            $35.2$ $(+5\%)$ & ${\bf 41.2}$ ${\bf (+24\%)^{\dagger}}$ &           $39.1$ $(+17\%)$ &           $38.8$ $(+16\%)$ \\
Relation. and Pers. Reflect. (198-224) & $38.1$ $(+14\%)$ &           $37.4$ $(+12\%)$ &       ${\bf 48.0}$ ${\bf (+44\%)^{*}}$ &           $41.8$ $(+26\%)$ &           $38.1$ $(+14\%)$ \\
                        Specific Info (537-554) & $38.9$ $(+17\%)$ &           $36.7$ $(+10\%)$ &            $35.2$ $(+6\%)$ &           $38.4$ $(+15\%)$ &       ${\bf 41.1}$ ${\bf (+23\%)^{*}}$ \\
                          Translation (173-197) & $38.9$ $(+17\%)$ &           $38.4$ $(+15\%)$ &            $35.9$ $(+8\%)$ &            $34.9$ $(+5\%)$ &           $40.5$ $(+22\%)$ \\
                 Tutoring or Teaching (431-464) & $38.7$ $(+16\%)$ & ${\bf 40.4}$ ${\bf (+21\%)^{\dagger}}$ &           $37.3$ $(+12\%)$ &           $39.0$ $(+17\%)$ &            $36.5$ $(+9\%)$ \\
                              Unclear (360-398) & $38.1$ $(+14\%)$ &           $32.4$ $(-3\%)$ &           $39.0$ $(+17\%)$ &            $34.2$ $(+3\%)$ &           $40.3$ $(+21\%)$ \\
                        Write Fiction (269-297) & $40.9$ $(+23\%)$ &           $38.0$ $(+14\%)$ &           $39.0$ $(+17\%)$ &           $38.2$ $(+15\%)$ &           $37.0$ $(+11\%)$ \\
\end{tabular}
  \begin{tablenotes}
    \item We used a Mcnemar test to compare model performance with the random classification baseline. Inferrences achieving a p-value $< 0.1$ are in bold and annotated as follow: $^{\dagger}p < 0.1$, *$p < 0.05$, **$p < 0.01$.
   \end{tablenotes}
 \end{threeparttable}
\end{table*}

\subsection{Performance Evolution over Number of Chats}

Finally, we analyzed how general inference varies with the number of chats a user has in their history. \autoref{table:acc_n_chats} depicts the inference accuracy for users depending on how many chats they have. We can see that for all traits except conscientiousness, inference for users with more than $100$ chats clearly outperforms inference for users with fewer chats. This stresses the fact that the more information is shared, the more information the model will have to proceed with the inference, especially with extraversion, agreeableness, and neuroticism.
\section{Discussion \& Conclusion}
Our experimental results show that user input in CAs brings valuable information to classify users according to all five personality traits. On the one hand, we could not achieve significant results with zero-shot inference, therefore showing the limitation of such method for data collected on the field, compared to similar work based on in-lab collection~\cite{wright_assessing_2026}. But on the other hand, we showed that fine-tuned models can perform significantly better than random in many cases. We achieved general single-user classification significantly better than random for agreeableness and neuroticism and showed that inference can be significantly achieved for all traits but conscientiousness (for which we only have weak evidence, i.e., $p<0.1$), depending on the type of data or the use case.
In general, the accuracy levels achieved in our experiments are close to previous similar work, e.g., $+20\%$ compared to random for extraversion~\cite{rapp_exploring_2024} with LLM-based models from essays, and other work with an accuracy of $+35.0\%$ to $61.8\%$ across all five traits, depending on the used dataset~\cite{zhu_lexical_2022}. Our method, however, depicts results for ternary classification instead of binary~\cite{demerdash_deep_2022,zhu_lexical_2022,rapp_exploring_2024}, thereby offering a finer assessment of the risks. Our results also suggest that the more a user interacts with CAs, the more it is likely for them to be correctly classified regarding their personality traits, therefore showing a growing threat over the degree of adoption(i.e., the more users there are, and the more they use), but also over time (i.e., the longer they use).

The limitations of our approach include focusing on English-speaking users and one specific CA (i.e., ChatGPT). Furthermore, we relied on online recruitment platforms for data collection~\cite{westwood_potential_2025}. Yet, our cautious methodology ensures trustworthy results in that our data collection method has two main advantages. (1) It guarantees that our models have no previous knowledge of the ground truth, which is not necessarily the case with models trained with public datasets, as LLMs are generally pre-trained on data scraped from the web. (2) The results are based on actual CA user data, and thus correspond to a realistic adversary model.

\begin{table*}
\centering
\caption{Inference accuracy across users having different number of chats in their history based on majority vote for each trait. Min. and max. number of users for each case are in parentheses.}
\label{table:acc_n_chats}
\begin{threeparttable}
\rowcolors{1}{white}{tblgray}

\begin{tabular}{l|l|l|l|l|l}
                         {\bf N chats} &                    {\bf O} &                          {\bf C} &                          {\bf E} &                    {\bf A} &                          {\bf N}\\
\midrule
       $<$20 (285-304) &  $34.8$ $(+4\%)$ &           $32.3$ $(-3\%)$ &           $37.4$ $(+12\%)$ &           $38.6$ $(+16\%)$ &           $38.5$ $(+15\%)$ \\
        20-100 (212-224) &  $39.3$ $(+18\%)$ &           $40.1$ $(+20\%)$ &           $36.7$ $(+10\%)$ &           $37.2$ $(+12\%)$ &           $39.7$ $(+19\%)$ \\
        100+ (132-163) & $42.4$ $(+27\%)$ &           $33.8$ $(+1\%)$ &           $48.4$ $(+45\%)$ &           $46.2$ $(+39\%)$ &            $44.8$ $(+34\%)$ \\
\end{tabular}
 \end{threeparttable}
\end{table*}

Moreover, we achieved our results from data of N=668 users, not all of which share the same type of data or have the same use cases. Given the current number of users of main CA platforms (e.g., $>800$M ChatGPT users in January 2026~\cite{duarte_number_2026}), we can consider our results as a lower bound for inference accuracy, as adversaries might have access to a much larger dataset. These adversaries not only include service providers interested in increasing their revenue through targeted advertisement, but also external malicious attackers getting access to user data with diverse methods (as described in Section \ref{section:threat}). Furthermore, some service providers and other potential adversaries have access to larger computational resources to use state-of-the-art LLMs that could perform better than our base model. 
Moreover, multiple large companies, such as Google, Meta, and X Corp., also own, in addition to CA/LLM-model-based services, other services and products, such as online social networks, search engines, or smartwatches. These provide them with additional user data to potentially proceed to cross-source inference.

Our results also depict privacy risks related to data-sharing behavior and usage patterns. Such fine-grained inference analysis allows us to assess which data types or use cases pose the most privacy risks related to personality traits inference. Our work not only brings valuable information for risk assessment, but also for helping the development and analysis of mitigation techniques, which should be tackled by future works. For example, our results could be used by researchers and developers to set up specific tools for lay users (e.g., language-model-based) to support them in identifying risky behavior and automatically alert them of sensitive data types in prompts. Beyond warnings, these tools could also provide mitigation strategies such as de-identifying or even removing certain data from prompts. Such a tool could, for example, be a locally-run text processing browser extension, or directly be implemented by the service provider that would enhance user privacy, especially if their services are based on API queries and are not directly managing models.

Based on our results, we identify moderate privacy risks for a specific individual and conclude that analyzing and inferring information from user-CA interactions could constitute major risks at large scale. Given that a few service providers concentrate a tremendous number of CA users, personal data, and, in particular, information about personality traits, could be leveraged for misuse cases ranging from targeted advertisement to large-scale manipulation campaigns spreading disinformation and/or political propaganda. Setting up mitigation techniques is thus crucial and should be a priority in AI research and R\&D.

\begin{credits}
\subsubsection{\ackname}
The research described in this paper was partially supported by the Hasler Foundation with Grant \#2024-04-29-51.
\end{credits}

\bibliographystyle{splncs04}
\bibliography{references}

\newpage
\appendix
\section{Dataset}

\subsection{User Chats}
\begin{table}[H]
\centering
\begin{filecontents*}{final_metadata/n_chats.csv}
total,mean,median,std,min,max
62090,92.81,28.0,185.72,1,2209
\end{filecontents*}
\csvautotabular{final_metadata/n_chats.csv}
\caption{Number of chats per user}
\end{table}

\subsection{User Input Format}
\begin{lstlisting}
Conversation Topic: [chat title]
- [user input 1]
- [user input 2]
...
\end{lstlisting}

\subsection{Label Thresholds}
IPIP-60 scores from the survey are scaled to the range 0-100. Each quantile holds one third of the scores.
\begin{verbatim}
"openness": [
            0.0,
            56.25,
            66.66666666666666,
            100.0
        ],
"conscientiousness": [
    0.0,
    64.58333333333334,
    77.08333333333334,
    100.0
],
"extraversion": [
    0.0,
    52.083333333333336,
    66.66666666666666,
    100.0
],
"agreeableness": [
    0.0,
    66.66666666666666,
    77.08333333333334,
    100.0
],
"neuroticism": [
    0.0,
    35.41666666666667,
    50.0,
    100.0
]
\end{verbatim}

\subsection{General Statistics}

\begin{table}[H]
\centering
\begin{filecontents*}{final_metadata/likert.csv}
trait,mean,std
openness,3.4517215568862274,0.5111695411809298
conscientiousness,3.8062624750498997,0.5466961507959252
extraversion,3.352295409181637,0.6633743446438441
agreeableness,3.830588822355289,0.521857692801121
neuroticism,2.741142714570858,0.7228875143565631
\end{filecontents*}
\csvautotabular{final_metadata/likert.csv}
\caption{Likert Scores for Personality Traits}
\end{table}

\begin{table}[H]
\centering
\begin{filecontents*}{final_metadata/country.csv}
,N
UK,341
USA,327
\end{filecontents*}
\csvautotabular{final_metadata/country.csv}
\caption{Countries}
\end{table}

\begin{table}[H]
\centering
\begin{filecontents*}{final_metadata/gender.csv}
,N
Man,358
Woman,303
Non-binary,4
Not disclosed,3
\end{filecontents*}
\csvautotabular{final_metadata/gender.csv}
\caption{Genders}
\end{table}

\subsection{UK/US Populations}

\begin{table}[H]
\centering
\begin{filecontents*}{final_metadata/likert_by_country.csv}
country,trait,mean,std
UK,openness,3.3726783968719456,0.494185665269059
UK,conscientiousness,3.8000977517106556,0.5317101434908015
UK,extraversion,3.3729227761485823,0.6752365556680416
UK,agreeableness,3.7421798631476046,0.555406380313245
UK,neuroticism,2.695992179863148,0.683216322792068
USA,openness,3.53414882772681,0.5162882190772518
USA,conscientiousness,3.812691131498471,0.5626431458361673
USA,extraversion,3.330784913353721,0.6511109186415823
USA,agreeableness,3.922782874617737,0.46773537730462966
USA,neuroticism,2.78822629969419,0.760256957351413
\end{filecontents*}
\csvautotabular{final_metadata/likert_by_country.csv}
\caption{Likert Scores for Personality Traits}
\end{table}

\subsection*{UK Label Thresholds}
\begin{verbatim}
"openness": [
    0.0,
    54.166666666666664,
    64.58333333333334,
    100.0
],
"conscientiousness": [
    0.0,
    64.58333333333334,
    77.08333333333334,
    100.0
],
"extraversion": [
    0.0,
    52.083333333333336,
    68.75,
    100.0
],
"agreeableness": [
    0.0,
    62.5,
    75.0,
    100.0
],
"neuroticism": [
    0.0,
    33.33333333333333,
    50.0,
    100.0
]
\end{verbatim}

\subsection*{US Label Thresholds}
\begin{verbatim}
"openness": [
    0.0,
    56.25,
    68.75,
    100.0
],
"conscientiousness": [
    0.0,
    64.58333333333334,
    77.08333333333334,
    100.0
],
"extraversion": [
    0.0,
    52.083333333333336,
    66.66666666666666,
    100.0
],
"agreeableness": [
    0.0,
    68.75,
    79.16666666666666,
    100.0
],
"neuroticism": [
    0.0,
    35.41666666666667,
    52.083333333333336,
    100.0
]
\end{verbatim}

\section{Experimental Setup}

\subsection{Hardware Setup}

\begin{itemize}
    \item \textbf{GPU} NVIDIA L40 GPU with the 16Q vGPU profile
    \item \textbf{CPU} Intel(R) Xeon(R) Gold 6442Y (4 cores)
    \item \textbf{RAM} 16GB
\end{itemize}

\subsection{User Chat Analysis}
\subsection*{Non-default Parameters}
\begin{verbatim}
{
    "Qwen/Qwen3-8B-AWQ (vLLM)": {
        "dtype": "auto",
        "max_model_len": 5120,
        "gpu_mem_util": 0.8,
        "tp": 1,
        "max_num_seqs": 4,
        "block_size": 16,
        "swap_space": 8,
        "seed": 42
    }
}
\end{verbatim}

\subsection*{Data Types}
\begin{table}[h]
\centering
\begin{filecontents*}{final_metadata/category_percentages_by_key_data_type.csv}
category,percentage
Identifiers and General Information,5.637406695133479
Lifestyle and Health,34.85609048534828
Personal Characteristics and Emotions,4.586199028651521
Standard of Living and Opinions,54.92030379086672
\end{filecontents*}
\csvautotabular{final_metadata/category_percentages_by_key_data_type.csv}
\caption{Portion of Chats in Overall Data Sharing Categories}
\end{table}

\begin{table}[h]
\centering
\begin{filecontents*}{final_metadata/category_percentages_data_type.csv}
category,percentage
Wealth Details Salary,7.135173897454285
Mental Health,1.6167410932559731
Family Life And Relationship,5.714006323543792
Recreational Consumption,6.8467029564197
Job And Education,38.80015645881548
Personal Identifiers,2.0844877603572476
Lifestyle And Habits,17.347371165944132
Physical Health Diagnosis,8.19127090191988
Physical Traits,0.6486521724958441
Political Views,1.3396786075165423
Gender,0.22979888523093975
Mental State And Personality Mood,3.083542488347078
Sexual And Dating Activities,0.8540043678085987
Location And Mobility Homeplace,2.560383324097917
Religion,1.9312885035366212
Account Credentials,0.42374262524854134
Age,0.4856742397079435
Sexual Orientation,0.13853124286971544
Ethnicity And Citizenship,0.4856742397079435
Criminal Records,0.08311874572182927
\end{filecontents*}
\csvautotabular{final_metadata/category_percentages_data_type.csv}
\caption{Portion of Chats in Data Sharing Sub-Categories}
\end{table}

\paragraph{Prompt}\par\noindent\par
\UseRawInputEncoding
\footnotesize
\begin{lstlisting}[breaklines=true]
You are a classifier that assigns a "Personal Data Type" to a set of user messages from a conversation with an AI assistant.

You will be given:
    - A Conversation Topic (chat title)
    - A list of User Messages (only the user’s inputs, no model responses)

**Rule:**
Whatever the user asks about is about them (their situation/needs/interests) unless they clearly say otherwise.

For example, if the user is asking about certain symptoms or drugs, they are likely asking because they have these symptoms or been prescribed this drug. Therefore, they are disclosing information about their health. If they're asking about investing strategies, that means they likely have funds to spare at the end of the month to be able to invest. Therefore, they are disclosing information about their salary and wealth. If they're asking about technical or scientific topics, it is likely because they the user has encountered these topics in their daily life, which most likely means that they came across these questions in their profession. If the user is asking about a TV show or a song, they're likely interested in it, therefore they are disclosing information about their lifestyle and habits, and so on.

Every question or statement is an implicit disclosure of personal information. Reason about how the user chat relates to a Personal Data Type.

Pick the single closest label; do not create a new one or choose "None". Assume the user always discloses some type of personal information, implicitly or explicitly.

Here are the Personal Data Type Choices:

- Personal Identifiers
- Account Credentials
- Location and Mobility Homeplace
- Ethnicity and Citizenship
- Criminal Records
- Mental Health
- Physical Health Diagnosis
- Lifestyle and Habits
- Recreational Consumption
- Sexual and Dating Activities
- Sexual orientation
- Mental State and Personality Mood
- Gender
- Age
- Physical Traits
- Family Life and Relationship
- Wealth Details Salary
- Job and Education
- Religion
- Political Views

Now classify the following chat.

{input}

Personal Data Type:
\end{lstlisting}

\subsection*{Use Case}
\begin{table}[h]
\centering
\begin{filecontents*}{final_metadata/category_percentages_by_key_user_intent.csv}
category,percentage
Writing,21.135651780813074
Practical Guidance,25.425425425425423
Technical Help,11.382350092027512
Multimedia,0.5699247634731506
Seeking Information,32.48086796473893
Self-Expression,3.434079240530853
Other/Unkown,5.571700732991055
\end{filecontents*}
\csvautotabular{final_metadata/category_percentages_by_key_user_intent.csv}
\caption{Portion of Chats in Overall Use Case Categories}
\end{table}

\begin{table}[h]
\centering
\begin{filecontents*}{final_metadata/category_percentages_user_intent.csv}
category,percentage
Purchasable Products,3.9523394362104036
Specific Info,25.549743291678777
"Health, Fitness, Beauty, Or Self-Care",4.249410701023605
Write Fiction,2.7027027027027026
Argument Or Summary Generation,5.224579418127806
Personal Writing Or Communication,5.566857179760405
How-To Advice,6.015693112467306
Creative Ideation,6.7874325938842075
Unclear,5.132551906745455
Relationships And Personal Reflection,1.9067454551325518
Generate Or Retrieve Other Media,0.5699247634731506
Tutoring Or Teaching,8.372889018050309
Games And Role Play,1.1995866834576512
Translation,1.5095740902192516
Mathematical Calculation,3.078885336949853
Edit Or Critique Provided Text,6.131938390002906
Computer Programming,6.994090865058607
Cooking And Recipes,2.978785236849753
Data Analysis,1.3093738900190512
Asking About The Model,0.4391488262456005
Greetings And Chitchat,0.3277471019406503
\end{filecontents*}
\csvautotabular{final_metadata/category_percentages_user_intent.csv}
\caption{Portion of Chats in Overall Use Case Categories}
\end{table}

\paragraph{Prompt}\par\noindent\par
\footnotesize
\begin{lstlisting}[breaklines=true][H]
You are a classifier that labels user conversations with an AI chatbot.

You will be given a conversation topic (chat title) and a list of only the user's messages from that chat (no assistant responses).

Your task is to read all the user messages together and decide which capability the user is clearly interested in, based on the full conversation.

- **Edit or Critique Provided Text**: Improving or modifying text provided by the user. Examples: "Please shorten this paragraph.", Here’s my draft speech; can you suggest enhancements?"
- **Argument or Summary Generation**: Creating arguments or summaries on topics notprovided in detail by the user. Examples: "Make an argument for why the national debt is important.", "Provide a summary of the theory of relativity."
- **Personal Writing or Communication**: Assisting with personal messages, emails, or social media posts. Examples: Write a nice birthday card note for my girlfriend.", Help me write a cover letter for a job application."
- **Write Fiction**: Crafting poems, stories, or fictional content. Examples: "Write a poem about the sunset.", Create a short story about a time-traveling astronaut."
- **How-to Advice**: Providing step-by-step instructions or guidance on how to perform tasks or learn new skills. Examples: "My car won’t start; what should I try?", "What’s the best way to clean hardwood floors?"
- **Creative Ideation**: Generating ideas or suggestions for creative projects or activities. Examples: "What should I talk about on my future podcast episodes?", "Brainstorm names for a new coffee shop."
- **Tutoring or Teaching**: Explaining concepts, teaching subjects, or helping the user understand educational material. Examples: "Can you explain derivatives and integrals?", "Explain the causes of the French Revolution."
- **Translation**: Translating text from one language to another. Examples: "How do you say Happy Birthday in Hindi?", "Translate I love coding to German."
- **Mathematical Calculation**: Solving math problems, performing calculations, or working with numerical data. Examples: "What is 400000 divided by 23?", "What’s the integral of sin(x)?", "Convert 150 kilometers to miles."
- **Computer Programming**: Writing code, debugging, explaining programming concepts, or discussing programming languages and tools. Examples: "How to group by and filter for biggest groups in SQL.", "Explain how inheritance works in Java."
- **Purchasable Products**: Inquiries about products or services available for purchase. Examples: "What’s the best streaming service?", Recommend a good laptop under $1000."
- **Cooking and Recipes**: Seeking recipes, cooking instructions, or culinary advice. Examples: "How to cook salmon.", "Is turkey bacon halal?", "Give me a step-by-step guide to make sushi."
- **Health, Fitness, Beauty, or Self-Care**: Seeking advice or information on physical health, fitness routines, beauty tips, or self-care practices. Examples: "How to do my eyebrows.", "How can I improve my cardio fitness?", "Give me tips for reducing stress."
- **Specific Info**: Providing specific information typically found on websites, including information about well-known individuals, current events, historical events, and other facts and knowledge. Examples: "What is regenerative agriculture?", "Tell me about Marie Curie and her main contributions to science."
- **Greetings and Chitchat**: Casual conversation, small talk, or friendly interactions without a specific informational goal. Examples: I had an awesome day today; how was yours?", "Whats your favorite animal?"
- **Relationships and Personal Reflection**: Discussing personal reflections or seeking advice on relationships and feelings. Examples: "what should I do for my 10th anniversary?", "My wife is mad at me, and I don’t know what to do."
- **Games and Role Play**: Engaging in interactive games, simulations, or imaginative role-playing scenarios. Examples: "You are a Klingon. Lets discuss the pros and cons of working with humans.", "I want you to be my AI girlfriend."
- **Asking About the Model**: Questions about the AI models capabilities or characteristics. Examples: "How many languages do you speak?", "Who made you?", "What do you know?"
- **Generate or Retrieve Other Media**: Creating or finding media other than text or images, such as audio, video, or multimedia files. Examples:"Make a YouTube video about goal kicks.", "Create a spreadsheet for mortgage payments.", "Write PPT slides for a tax law conference."
- **Data Analysis**: Performing statistical analysis, interpreting datasets, or extracting insights from data. Examples: "Heres a spreadsheet with my expenses; tell me how much I spent on which categories."
- **Unclear**: If the user’s intent is not clear from the conversation. [If there is no indication of what the user wants; usually this would be a very short prompt.]"

Only reply with one of the capabilities above, without quotes and as presented. Choose the MOST SPECIFIC RELEVANT option.
If the conversation has multiple distinct capabilities, choose the one that is the most relevant to the **LAST message** in the conversation.

The response MUST follow this exact format:

User Intent: <one capability from the list above>

Now classify the following chat.

{input}

\end{lstlisting}

\subsection{Classifier Fine-tuning}
\subsection*{Non-default Parameters}
\begin{verbatim}
"FacebookAI/roberta-base": {
    "dataloader": {
        "per_device_train_batch_size": 128,
        "per_device_eval_batch_size": 128,
        "gradient_accumulation_steps": 1
    },
    "training": {
        "seed": 42,
        "bf16": true,
        "gradient_checkpointing": true,
        "warmup_ratio": 0.03,
        "weight_decay": 0.01,
        "optim": "adamw_torch",
        "metric_for_best_model": "accuracy",
        "greater_is_better": true,
        "load_best_model_at_end": true,
        "remove_unused_columns": false,
        "save_safetensors": true
    },
    "evaluation_and_saving": {
        "eval_strategy": "steps",
        "eval_steps": 85,
        "save_strategy": "steps",
        "save_steps": 85,
        "save_total_limit": 8
    }
}
\end{verbatim}

\subsection*{K-fold Cross Validation}

\begin{table}[H]
  \centering
  \begin{filecontents*}{final_metadata/overall_fold_stats.csv}
trait,K,N,n_unique_RID,unique_rids_per_fold_mean,unique_rids_per_fold_std,fold_size_mean,fold_size_std,label0_mean,label1_mean,label2_mean
O,5,46432,655,131,17.53,9286.4,383.79,33.13,33.9,32.98
C,5,49618,657,131.4,19.19,9923.6,334.41,33.34,33.24,33.42
E,5,60121,668,133.6,19.03,12024.2,958.67,33.26,33.59,33.16
A,5,58607,667,133.4,8.62,11721.4,867.7,33.6,33.51,32.9
N,5,49672,659,131.8,4.96,9934.4,418.64,33.35,33.05,33.6
\end{filecontents*}
\csvreader[
    tabular=llllrrrrrrr,
    table head=\toprule
      Trait & $K$ & Total $N$ & Total \# RIDs &
      RIDs (\Mean) &
      RIDs (\Std) &
      Fold (\Mean) &
      Fold (\Std) &
      Lbl 0 (\Mean) &
      Lbl 1 (\Mean) &
      Lbl 2 (\Mean) \\
      \midrule,
    late after line=\\,
    table foot=\bottomrule
  ]{final_metadata/overall_fold_stats.csv}{%
    trait=\Trait,
    K=\K,
    N=\N,
    n_unique_RID=\NumRID,
    unique_rids_per_fold_mean=\URIDMean,
    unique_rids_per_fold_std=\URIDStd,
    fold_size_mean=\FoldMean,
    fold_size_std=\FoldStd,
    label0_mean=\Lzero,
    label1_mean=\Lone,
    label2_mean=\Ltwo
  }{
    \Trait & \K & \N & \NumRID &
    \URIDMean & \URIDStd &
    \FoldMean & \FoldStd &
    \Lzero & \Lone & \Ltwo
  }
\end{table}

\subsection*{Fold Sizes}

\begin{table}[H]
\centering
\begin{filecontents*}{final_metadata/fold_lengths.csv}
trait,fold0,fold1,fold2,fold3,fold4
openness,8992,9421,9217,8853,9949
conscientiousness,10058,10377,10082,9424,9677
extraversion,11919,10870,12334,13662,11336
agreeableness,11349,11194,13305,11930,10829
neuroticism,9994,9382,10423,9531,10342
\end{filecontents*}
\csvautotabular{final_metadata/fold_lengths.csv}
\caption{Number of data points per fold}
\end{table}

\subsection*{Grid Search Parameters}

\begin{verbatim}
{
  "learning_rates": [3e-5, 5e-5, 1e-4, 2e-4],
  "lr_scheduler": ["linear", "cosine"],
  "early_stopping_patiences": [6, 8],
  "early_stopping_thresholds": [0.001, 0.002]
}
\end{verbatim}

\subsection*{Grid Search Results}

\begin{table}[H]
\centering
\begin{filecontents*}{final_results/openness_eval_accuracies.csv}
lr scheduler,learning rate,early stopping patience,early stopping threshold,mean eval acc
linear,0.0001,8,0.001,0.423442799196962
cosine,0.0001,6,0.002,0.423291457909085
cosine,0.0001,6,0.001,0.422794284680923
cosine,0.0001,8,0.002,0.419741779971101
linear,0.0001,8,0.002,0.418253978341017
linear,0.0001,6,0.001,0.415270550515183
linear,0.0002,6,0.002,0.414542378487724
linear,0.00005,8,0.001,0.413691596726242
cosine,0.0001,8,0.001,0.412154132444252
cosine,0.0002,8,0.002,0.410775207377326
linear,0.0002,8,0.001,0.410525493792285
linear,0.0001,6,0.002,0.41046220809824
cosine,0.0002,6,0.001,0.408913435284661
linear,0.0002,6,0.001,0.408009103917434
cosine,0.00003,8,0.002,0.407893620507422
cosine,0.00005,8,0.001,0.407683011019232
cosine,0.0002,8,0.001,0.407583069838148
cosine,0.00005,8,0.002,0.406606683299694
linear,0.00005,6,0.002,0.406040862730308
cosine,0.00005,6,0.002,0.40496599019044
cosine,0.00003,6,0.002,0.404659156082335
cosine,0.0002,6,0.002,0.404347227911742
cosine,0.00003,6,0.001,0.403866535843476
linear,0.00005,6,0.001,0.403049335186545
linear,0.00003,8,0.001,0.402912915347034
cosine,0.00003,8,0.001,0.402909068622099
linear,0.00003,6,0.001,0.402000387064328
linear,0.00005,8,0.002,0.400885181900278
cosine,0.00005,6,0.001,0.400248991385416
linear,0.0002,8,0.002,0.398909475330418
linear,0.00003,8,0.002,0.398344432543322
linear,0.00003,6,0.002,0.398288620714895
\end{filecontents*}
\csvautotabular{final_results/openness_eval_accuracies.csv}
\caption{Openness Grid Search Results}
\end{table}

\begin{table}[H]
\centering
\begin{filecontents*}{final_results/conscientiousness_eval_accuracies.csv}
lr scheduler,learning rate,early stopping patience,early stopping threshold,mean eval acc
linear,0.00005,8,0.001,0.456964732486518
linear,0.00005,6,0.001,0.45635668557433
cosine,0.00005,6,0.002,0.455880799931423
linear,0.0001,8,0.002,0.455296704891737
linear,0.00003,8,0.001,0.454701145889551
cosine,0.00005,8,0.002,0.454431013098535
cosine,0.00005,6,0.001,0.453360356815939
linear,0.00005,6,0.002,0.453327928280349
linear,0.00003,6,0.002,0.452836755720713
cosine,0.00003,8,0.002,0.452461842760635
cosine,0.00005,8,0.001,0.452434639355978
cosine,0.00003,6,0.001,0.452092274038606
cosine,0.00003,6,0.002,0.450836514548004
linear,0.00003,8,0.002,0.449686096583238
linear,0.0001,6,0.001,0.449375825960516
cosine,0.00003,8,0.001,0.447756507897741
linear,0.00005,8,0.002,0.44761631567439
cosine,0.0001,8,0.002,0.447272421650911
cosine,0.0001,6,0.002,0.446587836165661
cosine,0.0001,8,0.001,0.445414310723382
cosine,0.0002,8,0.001,0.444769986803016
cosine,0.0002,6,0.001,0.444265573616993
linear,0.0001,8,0.001,0.444186684945661
cosine,0.0001,6,0.001,0.443044150038572
linear,0.0001,6,0.002,0.442764903424256
cosine,0.0002,8,0.002,0.440135592451764
linear,0.00003,6,0.001,0.439830962047442
linear,0.0002,8,0.002,0.438064880983765
linear,0.0002,6,0.002,0.437021354210223
linear,0.0002,8,0.001,0.435116387442683
linear,0.0002,6,0.001,0.423052189337465
cosine,0.0002,6,0.002,0.422004574928886
\end{filecontents*}
\csvautotabular{final_results/conscientiousness_eval_accuracies.csv}
\caption{Conscientiousness Grid Search Results}
\end{table}

\begin{table}[H]
\centering
\begin{filecontents*}{final_results/extraversion_eval_accuracies.csv}
lr scheduler,learning rate,early stopping patience,early stopping threshold,mean eval acc
cosine,0.0002,8,0.002,0.463173779952542
linear,0.0001,8,0.002,0.461384401545008
cosine,0.00003,8,0.001,0.459052073464082
cosine,0.00007,8,0.001,0.457501884550442
linear,0.0001,6,0.001,0.45704788752954
linear,0.00005,8,0.002,0.45654346831378
linear,0.0002,6,0.001,0.455808118195182
cosine,0.00007,6,0.002,0.455334851754686
linear,0.0001,8,0.001,0.454977808123829
cosine,0.0001,8,0.002,0.454535639798237
cosine,0.00005,8,0.001,0.454029207511255
cosine,0.00003,6,0.001,0.453917660937297
linear,0.00005,6,0.001,0.453795138480558
linear,0.0001,6,0.002,0.452679523579426
linear,0.00003,8,0.002,0.452102770548828
cosine,0.0001,6,0.002,0.451360384553227
linear,0.00005,8,0.001,0.45131572382823
linear,0.00003,6,0.001,0.451047812608795
cosine,0.00007,6,0.001,0.450635058077548
cosine,0.00005,8,0.002,0.449696145342455
cosine,0.00003,6,0.002,0.449437710999396
cosine,0.00005,6,0.001,0.449322854914685
cosine,0.0002,8,0.001,0.449215787699758
linear,0.00003,6,0.002,0.448942837863196
linear,0.0002,8,0.002,0.448833741409615
cosine,0.0001,8,0.001,0.448094952798663
cosine,0.00003,8,0.002,0.447417826322204
cosine,0.00007,8,0.002,0.447302447617308
cosine,0.0001,6,0.001,0.445486891558081
cosine,0.0002,6,0.001,0.445244779105255
linear,0.00003,8,0.001,0.444463608479216
linear,0.0002,6,0.002,0.444030710357681
cosine,0.00005,6,0.002,0.443270480466977
linear,0.00005,6,0.002,0.44268060174929
linear,0.0002,8,0.001,0.441854031827596
cosine,0.0002,6,0.002,0.4414142343464
\end{filecontents*}
\csvautotabular{final_results/extraversion_eval_accuracies.csv}
\caption{Extraversion Grid Search Results}
\end{table}

\begin{table}[H]
\centering
\begin{filecontents*}{final_results/agreeableness_eval_accuracies.csv}
lr scheduler,learning rate,early stopping patience,early stopping threshold,mean eval acc
cosine,0.0001,8,0.002,0.438769577960131
cosine,0.00003,8,0.002,0.434040767017007
cosine,0.00005,8,0.002,0.433982076973121
linear,0.00003,8,0.001,0.433716967593989
cosine,0.0001,6,0.002,0.432876760077671
linear,0.0001,8,0.002,0.432867030362718
linear,0.0001,6,0.002,0.431798800787397
cosine,0.00005,6,0.002,0.431611648457329
cosine,0.00003,8,0.001,0.431488754009141
cosine,0.00003,6,0.002,0.431436327329881
linear,0.00003,6,0.001,0.431302415535779
cosine,0.00003,6,0.001,0.431205157152788
cosine,0.0002,8,0.002,0.431143253123054
linear,0.00005,8,0.002,0.431113536072934
cosine,0.00005,6,0.001,0.430920605581115
linear,0.0002,6,0.002,0.430431201354836
linear,0.00005,8,0.001,0.429985147030862
linear,0.00005,6,0.001,0.429918772715825
linear,0.00005,6,0.002,0.429241941879083
linear,0.00003,6,0.002,0.429032452237514
cosine,0.0002,6,0.002,0.428488827245797
linear,0.00003,8,0.002,0.428208580949137
cosine,0.0001,6,0.001,0.42817661832486
cosine,0.0001,8,0.001,0.427010991713217
cosine,0.00005,8,0.001,0.425946288380979
linear,0.0001,6,0.001,0.425350808857899
linear,0.0002,6,0.001,0.42531258063896
linear,0.0001,8,0.001,0.425267870067921
linear,0.0002,8,0.002,0.420846040809068
cosine,0.0002,8,0.001,0.419176157745659
cosine,0.0002,6,0.001,0.418759018067555
linear,0.0002,8,0.001,0.417683125289181
\end{filecontents*}
\csvautotabular{final_results/agreeableness_eval_accuracies.csv}
\caption{Agreeableness Grid Search Results}
\end{table}

\begin{table}[H]
\centering
\begin{filecontents*}{final_results/neuroticism_eval_accuracies.csv}
lr scheduler,learning rate,early stopping patience,early stopping threshold,mean eval acc
cosine,0.0001,8,0.001,0.417733213489891
linear,0.0001,8,0.001,0.412928474273553
cosine,0.00003,8,0.002,0.412897222000428
linear,0.00003,8,0.001,0.411254357170752
cosine,0.00003,8,0.001,0.411178855232779
cosine,0.00005,6,0.001,0.410889497727153
linear,0.00005,6,0.001,0.4097775395078
linear,0.00005,8,0.002,0.409646980186065
cosine,0.0001,8,0.002,0.409484482556087
linear,0.0002,8,0.001,0.409227601937941
linear,0.00005,8,0.001,0.408487048953111
cosine,0.0002,8,0.002,0.408049128544185
cosine,0.0001,6,0.001,0.407641294839283
cosine,0.0002,6,0.002,0.407151119022945
linear,0.00003,8,0.002,0.406367679791089
cosine,0.00005,8,0.001,0.405883571528406
cosine,0.00003,6,0.002,0.404496778648071
linear,0.00003,6,0.002,0.404107124783239
cosine,0.00005,8,0.002,0.403519512786821
linear,0.00005,6,0.002,0.403416550180942
cosine,0.00005,6,0.002,0.402647275438991
linear,0.0001,6,0.001,0.402631749919278
cosine,0.00003,6,0.001,0.402220966802065
linear,0.0002,8,0.002,0.402063408699181
cosine,0.0002,6,0.001,0.401508007799129
cosine,0.0001,6,0.002,0.400934898162137
linear,0.0002,6,0.001,0.400090947615753
linear,0.0001,6,0.002,0.399880868923782
cosine,0.0002,8,0.001,0.399487865752057
linear,0.00003,6,0.001,0.399160478849963
linear,0.0001,8,0.002,0.39835688291451
linear,0.0002,6,0.002,0.396775247886052
\end{filecontents*}
\csvautotabular{final_results/neuroticism_eval_accuracies.csv}
\caption{Neuroticism Grid Search Results}
\end{table}

\end{document}